# Automated Fidelity Assessment for Strategy Training in Inpatient Rehabilitation using Natural Language Processing

Hunter Osterhoudt[1], Courtney E. Schneider, OTD, OT[2], Haneef A Mohammad[3], Minmei Shih, PhD, OTR/L[2], Alexandra E. Harper, MOT, OTR/L[2] Leming Zhou, PhD[4], Elizabeth R Skidmore, PhD, OTR/L[2], Yanshan Wang, PhD[4,5,6]

[1]Department of Computer Science, School of Computing and Information, University of Pittsburgh; [2]Department of Occupational Therapy, School of Health and Rehabilitation Sciences, University of Pittsburgh; [3]Department of Information Science, School of Computing and Information, University of Pittsburgh; [4]Department of Health Information Management, School of Health and Rehabilitation Sciences, University of Pittsburgh; [5]Department of Biomedical Informatics, University of Pittsburgh, Pittsburgh, PA; [6]Intelligent Systems Program, University of Pittsburgh, Pittsburgh, PA

**Abstract**
*Strategy training is a multidisciplinary rehabilitation approach that teaches skills to reduce disability among those with cognitive impairments following a stroke. Strategy training has been shown in randomized, controlled clinical trials to be a more feasible and efficacious intervention for promoting independence than traditional rehabilitation approaches. A standardized fidelity assessment is used to measure adherence to treatment principles by examining guided and directed verbal cues in video recordings of rehabilitation sessions. Although the fidelity assessment for detecting guided and directed verbal cues is valid and feasible for single-site studies, it can become labor intensive, time consuming, and expensive in large, multi-site pragmatic trials. To address this challenge to widespread strategy training implementation, we leveraged natural language processing (NLP) techniques to automate the strategy training fidelity assessment, i.e., to automatically identify guided and directed verbal cues from video recordings of rehabilitation sessions. We developed a rule-based NLP algorithm, a long-short term memory (LSTM) model, and a bidirectional encoder representation from transformers (BERT) model for this task. The best performance was achieved by the BERT model with a 0.8075 F1-score. This BERT model was verified on an external validation dataset collected from a separate major regional health system and achieved an F1 score of 0.8259, which shows that the BERT model generalizes well. The findings from this study hold widespread promise in psychology and rehabilitation intervention research and practice.*

**Introduction**
Stroke is a leading cause of disability in the United States. Meta-cognitive strategy training (henceforth referred to as strategy training) is a multidisciplinary rehabilitation approach that teaches skills to reduce disability among those with cognitive impairments following a stroke. In strategy training, the therapist stimulates the patient's learning through guided cueing in the form of prompts and questions rather than directed cueing or directed instructions.[1] Patients learn to identify and work through specific problems and everyday activities by using a meta-cognitive strategy that can promote an increased awareness of impaired skills as well as develop skills in self-assessment and plan formation to address areas of impairment. These skills can be applied to a variety of tasks and situations[1] and improve the ability to perform desired activities long after rehabilitation is completed.[2] Randomized, controlled clinical trials have demonstrated that strategy training is a feasible and more efficacious approach for promoting patient independence than traditional rehabilitation approaches.[1]

A cornerstone of strategy training is the emphasis on guided cueing over directed cueing. In traditional rehabilitation care, therapists typically identify barriers, solve problems, and instruct patients in solutions to improve patient performance of everyday activities.[2] They use directed verbal cues that are composed of therapist-initiated instructional statements and commands to elicit specific behavior (e.g., "We are going to work on this today," "Do it this way," etc.). Directed cueing improves performance on trained tasks but is less likely to result in skills that generalize to different contexts or tasks. For example, if even though a patient may be trained to master toilet transfers in an inpatient rehabilitation facility, the performance may not carry over to home after the hospital discharge due to differences in the environment.[3] Conversely, guided cueing facilitates the learning ability of patients to discover a strategy to solve problems.[2] Guided verbal cues are open-ended questions or statements that help patients to reflect on and evaluate performance (e.g., "What do you think went well?") and allow patients to discover and plan a strategy

(e.g., "How might you address that problem?" "Walk me through your plan."). When incorporated in strategy training, clinical trials demonstrate that guided cueing results in significantly greater reductions in disability compared to directed cueing.[4]

At present, treatment fidelity of strategy training in general, and guided cueing in particular, is evaluated using a standardized fidelity assessment.[5,6] This approach has been proven feasible and valid for single-site studies. However, the current approach requires a trained, independent evaluator (e.g., does not provide treatment in either strategy training sessions or usual care sessions) to manually review recorded rehabilitation session videos.[6] Trained evaluators are costly and the method for fidelity assessment is time-consuming, since it requires thorough attention to detail.[5] As the strategy training is being implemented across multiple sites, the current fidelity assessment method is infeasible. To address this limitation, we will leverage natural language processing (NLP) techniques to automate the process of strategy training fidelity assessment. Using automated speech recognition and text analysis, we plan to create an NLP system that can identify and extract guided and directed verbal cues from the transcripts of the recorded rehabilitation session videos. In doing so, we will develop a rule-based NLP system as well as machine learning- and deep learning-based NLP systems for automated fidelity assessment and examine which method will achieve better performance.

**Related Work**
Our team developed and examined the feasibility and reliability of the standardized fidelity assessment.[5,6] The standardized fidelity assessment characterizes the types (guided, directed), frequencies, and duration of training cues, and was derived through consensus review of multiple strategy training sessions by experts in the field. On average, the current approach of standardized fidelity assessment takes 1 – 2 minutes for every 1 minute of recorded video. Given the cost of a trained evaluator, this method costs between $40 to $100 for every 30-minute rehabilitation session. Most patients receive between 10 to 20 strategy training sessions per inpatient rehabilitation stay. As our team is planning a 30-site pragmatic trial examining implementation and adoption of strategy training in inpatient rehabilitation facilities across the nation, the manual fidelity assessment becomes labor intensive, costly, and infeasible.

Only recently have the fields of psychology and rehabilitation begun to characterize intervention elements in a standardized fashion with a view toward future efforts in standardized fidelity assessment.[7] Up to this point, the methods used to assess strategy training fidelity have been the state-of-the-art. However, to our knowledge, all are plagued with the same challenges in labor, time, and cost. The development of an automated method for standardized fidelity assessment is quite innovative and could have widespread implications when combined with current efforts to characterize and standardize psychological and rehabilitation interventions.

In recent years, due to the advancement in deep learning models, artificial intelligence (AI) techniques have become a viable option for different applications.[8-10] Automatic speech recognition (ASR) is an AI technique that involves recognition and translation of spoken language into text by computer algorithms.[11-13] Natural language processing (NLP), or computational linguistics, is another widely used AI technology that aims to analyze the text and extract useful information.[14-16] ASR in combination with NLP techniques have been applied in many healthcare applications. For example, ASR and NLP have been used to assess the language change in patients with mild cognitive impairment and Alzheimer's dementia[17]; and to identify suicidal behaviors and mental disorders[18]. To the best of our knowledge, ASR and NLP technologies have not been investigated to automate the fidelity assessment for strategy training in inpatient rehabilitation.

**Materials**
The recorded videos were generated from a clinical study conducted between 2019 and 2021, collaborating with a local inpatient rehabilitation facility in Pittsburgh, PA. The clinical study examined the feasibility of implementing strategy training in the flow of inpatient rehabilitation by a single multidisciplinary team of rehabilitation therapists. The research team observed traditional rehabilitation practices for three months and recorded 20% of intervention sessions for three disciplines, including occupational therapy (OT), physical therapy (PT), speech and language therapy (SLP). The research team then trained the clinical team in strategy training and repeated observations and recordings of 20% of sessions for 3 additional months. We collected a total of 100 video recordings representing 100 rehabilitation sessions, specifically OT (n=40), PT (n=38), and SLP (n=22), from 36 participants. We randomly selected 10 videos from each rehabilitation discipline with a total of 30 videos from the pool of 100 videos to be annotated for developing the NLP algorithm. For the 10 videos chosen from each discipline, 7 were randomly selected for training (with a total of 21 rehabilitation session videos for training), while the remaining 3 videos were used for validation (with a total of 9 rehabilitation session videos for validation).

**Methods**

We first created transcripts from the video recordings of the rehabilitation sessions. This was done using Amazon Web Service's ASR software. The software takes a video recording and returns a transcript of the media. Once received, we then cleaned the transcripts by removing unnecessary and redundant texts and punctuations to make them readable. The cleaned transcripts chosen were then sent to trained fidelity assessment evaluators for annotation. A previously trained evaluator trained a health informatics student to be an additional fidelity assessment evaluator. The two annotators were directed to annotate mentions of directed and guided verbal cues in the 30 video transcripts. We used the Multi-document Annotation Environment (MAE) tool for the annotation and create the gold standard dataset for developing NLP algorithms.

Initially, a batch of 3 documents was given to the annotators to refine the annotation guidelines and discuss discrepancies in order to reach a consensus on the concept definition. The annotators continued this process on another batch of 3 documents until the new annotator developed an understanding of all rehabilitation disciplines by achieving an inter-annotator agreement (IAA) above 0.70 in terms of Krippendorff's alpha in all three rehabilitation disciplines: OT, PT, and SLP. A third senior investigator resolved disagreements between the two annotators during this process. After the training process, the remaining transcription documents were split between the two annotators for independent annotation using the standardized fidelity assessment guidelines. Finally, we aggregated the directed and guided verbal cue annotations in a single document for training NLP algorithms. In order to train the proposed NLP algorithms, we also added a balanced number of sentences without any cues (i.e., none cues) so that we could train an NLP system to be able to distinguish between instructional cues and ordinary dialogue. The training dataset contains 784 guided verbal cues, 784 directed verbal cues, and 784 none cues, and the validation dataset contains 292 guided verbal cues, 289 directed verbal cues, and 290 none cues.

We developed a rule-based NLP algorithm and two deep learning-based NLP models to identify guided and directed verbal cues from the transcripts. For the rule-based NLP algorithm, we adopted data-driven and expert-driven approaches to create the fidelity assessment lexicon used to build the rules to automatically detect guided and directed verbal cues from rehabilitation sessions. In the data-driven approach, we used the annotated guided and directed verbal cues in the training data to develop the lexicon and the NLP rules. The NLP rules are the common trends found for a label based on the annotated transcripts. For example, many expressions in the lexicon that begin with interrogative pronouns (e.g., "what do you think," "how did you do," etc.) are guided, since many guided cues are questions framed in a way to stimulate thinking and problem solving. Figure 1 shows the process that we employed for the data-driven approach.

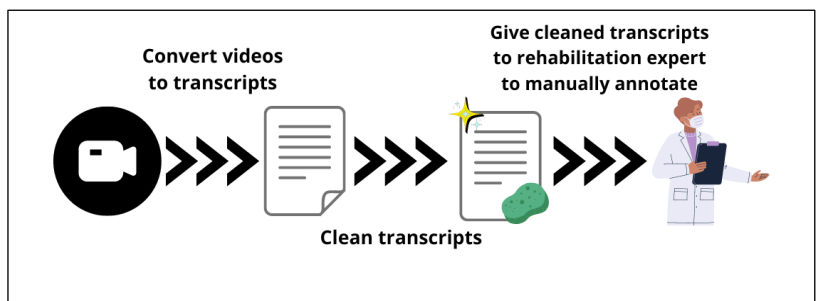

**Figure 1.** Pipeline and process of the data-driven approach

In the expert-driven approach, we leveraged the validated coding scheme used by the trained evaluators which includes common phrases and keywords of guided and directed verbal cues. Additionally, we consulted the evaluators on prevalent phrases and sentence structures that they have identified while annotating rehabilitation transcripts. Figure 2 shows the process employed for the expert-driven approach.

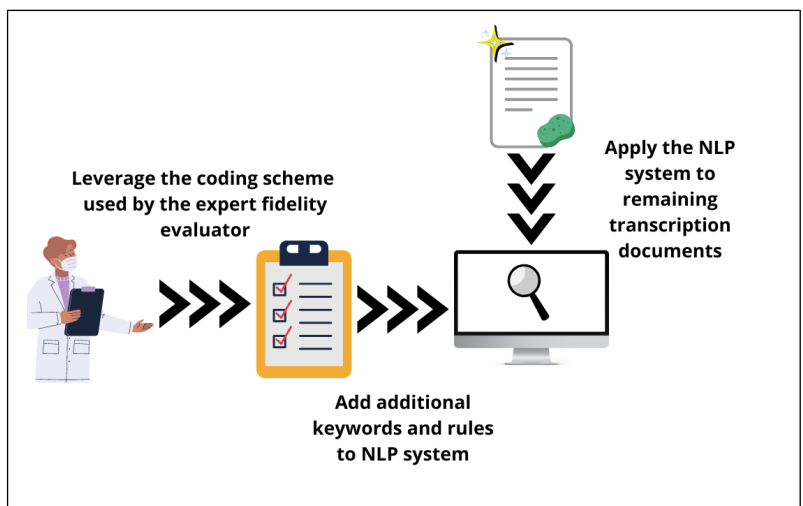

**Figure 2.** Pipeline and process of the expert-driven approach

Finally, using both the data-driven and expert-driven approaches, we developed a novel list of lexicons for fidelity assessment and then developed the rule-based NLP algorithm using regular expressions. **Table 1** exemplifies the format and structure of our fidelity assessment lexicon used to build the regular expressions dictionary. We compiled the lexicon into a list of regular expressions accompanied by the cue type that the statement was identified as (e.g., guided, directed, or none). This list was then used to search for and annotate matching cues within the gold standard transcripts. This model served as a means of validating our lexicon for identifying different instructional cues without employing more complex machine learning and NLP methods.

**Table 1.** Selected trigger words and examples from the fidelity assessment lexicon.

| Trigger Words | Examples | Cue Type |
| --- | --- | --- |
| do you | Do you want to write it out? | Guided |
|  | Do you need a drink of water? | None |
| if you | If you hear it a second time what happens… | Guided |
|  | If you give me two minutes I'm gonna run to the restroom… | None |
| look | Look for kind of those important landmarks in the room… <br> Look at that first one for me again… | Directed |
|  | Look better… | None |
| can you | Can think of any equipment that might help? <br> Can do that for me? <br> Can you see them? | Guided |
|  | Can you say those words backwards? <br> Can you keep your feet up? | Directed |
|  | Can pass it over to his left… | None |

**Table 1** (continued).

| what | What am I going to do next? | Guided |
| --- | --- | --- |
| | What feels sloppy… | |
| | What I'm gonna have you do you just work on a laundry basket in the kitchen… | Directed |
| | What I want you to do is come forward when you stand… | |
| | What if your feet get out? | None |
| | What about me could I have one…? | |
| let's | Let's talk about how you're gonna do it specifically… | Guided |
| | Let's come up with a plan for that… | |
| | Let's start… | Directed |
| | Let's go over to that area… | |
| | Let's try that again… | |
| | Let's give it a shot… | |

For the deep learning-based NLP algorithm, we trained a long-short term memory (LSTM) model and a bidirectional encoder representation from transformers (BERT) model to identify the guided and directed verbal cues. For the LSTM model, we used TensorFlow's Keras API to build a model with an embedding layer, LSTM with 100 neurons, and a dense layer with a sigmoid activation function using the Adam optimization algorithm and categorical cross entropy for loss calculation. For the BERT model, we trained a 12-layer base uncased BERT for Sequence Classification model from the Hugging Face Transformers Python package, using the Adam optimization algorithm and the provided binary cross entropy with logits loss equation for calculating loss during training.

The task was formulated as a text classification task where we classify a given sentence from the transcript into three categories: guided verbal cue, directed verbal cue, or none cue. We trained the LSTM model over 4 epochs with a batch size of 64, and the BERT model for 2 epochs with a batch size of 32.

To format the cues for training and testing the models, we removed leading, trailing, and double spaces. We then tokenized the text using Keras tokenizer for the LSTM model and BERT tokenizer for the BERT model. We set the maximum sequence length of our text to 64 since over 75% of our input data is less than 16 words in length, with a maximum length of 271 words. We found that the longer word counts are attributed to cues with the none label, which contain large sections of inconsequential dialogue irrelevant to the strategy training intervention. Finally, we truncate and/or pad the input sequences to our maximum sequence length.

Finally, we verify our best performing model using an external validation dataset consisting of 18 videos of strategy training rehabilitation sessions from a prior study. This dataset includes sessions with a different set of therapists who received the same protocol trainings but practiced in another major regional health system, and the recruited patients were with distinct cognitive impairments after stroke. We hypothesize that this will introduce variation in the language used since therapists may have a different way of speaking to patients. This testing dataset contains 274 guided verbal cues, 242 directed verbal cues, and 450 none cues.

**Results**
During the annotation process, the new evaluator experienced difficulty with the annotation of directed cues when compared to guided cues across all rehabilitation disciplines. The new evaluator found annotating OT sessions the most challenging when compared to other rehabilitation disciplines. This was attributed to the annotator's low understanding of the nature of the occupational therapy discipline. The expert annotator observed that the new annotator developed an understanding of SLP sessions the quickest and found it easier to code when compared to other rehabilitation disciplines. The IAA scores for OT, PT, SLP session transcripts during the annotation consensus phase were 0.73, 0.74, 0.79, respectively.

**Table 2** shows the training and validation statistics for the LSTM and BERT models. From these statistics, we can see that the LSTM model seems to overfit to the training data after the second training epoch. The LSTM and BERT models have both been optimized through hyperparameter tuning, and the results shown reflect the best performance that could be achieved by each of these models through experimentation. Both models were optimized by tuning the number of epochs and the batch size. The LSTM model was experimented on using different numbers of neurons, adding a dropout layer with a 0.2 dropout rate, and changing the activation function. The BERT model was optimized by setting the maximum sequence length appropriately, changing the type of optimizer, and changing the learning rate of the optimizer.

**Table 2.** Training and validation loss and accuracy for LSTM and BERT models

| Epoch | LSTM | | | | BERT | |
|---|---|---|---|---|---|---|
| | Training Loss | Training Accuracy | Validation Loss | Validation Accuracy | Average Training Loss | Validation Accuracy |
| 1 | 0.6773 | 0.4827 | 0.8478 | 0.6125 | 0.81 | 0.77 |
| 2 | 0.4602 | 0.7316 | 0.6173 | 0.7453 | 0.54 | 0.79 |
| 3 | 0.3419 | 0.7972 | 0.5833 | 0.7778 | | |
| 4 | 0.2886 | 0.8228 | 0.5983 | 0.7588 | | |

**Table 3** shows the F1 score for each of the NLP methods based on rehabilitation discipline and average of disciplines. To calculate the F1 score of each of our models, we used the scikit-learn metrics Python package's f1_score function for the multi-class (i.e., using 3 labels) case. Overall, the BERT model achieved the best performance on average with an F1-score of **0.8075** and the best performance for every rehabilitation discipline. Among three rehabilitation disciplines, the BERT model had the highest F1-score of 0.8261 for OT. The second-best method is the rule-based NLP model that achieved an average F1-score of 0.7058. The LSTM model underperformed compared to the other models even with extensive hyperparameter tuning. The reason might be due to the small size of training data. In contrast to the LSTM model, the BERT model takes advantage of the pre-training language model that learns language characteristics from the large corpora.

**Table 3.** F1 score of each model based on different rehabilitation disciplines.

| F1-score | Rule-based NLP | LSTM | BERT |
|---|---|---|---|
| OT | 0.7002 | 0.5507 | 0.8261 |
| PT | 0.6803 | 0.5549 | 0.7804 |
| SLP | 0.7301 | 0.5663 | 0.8100 |
| Average | 0.7058 | 0.5584 | 0.8075 |

Although it achieved the second-best performance results, developing the rule-based NLP system requires tedious effort to manually engineer rules and lexicons, which is very time-consuming. In addition, the rule-based NLP system suffered from one major drawback: the inability to consider context when classifying an example phrase. Take for instance the phrase "what do you think," which the evaluators did not mark as either a guided or directed cue in the training set. When we applied the regular expressions to the validation lexicon, whenever there was an example of a phrase containing "what do you think" it would then likely be marked as not a cue. Another example of this is the expression "if you need help". Given the context, this short phrase could be part of a guided or directed cue, or neither.

Since the phrase "if you need help" is a directed cue in the training set, any phrase that was identified using this regular expression would be marked as a directed cue regardless of its true label.

The LSTM model achieved the lowest performance results, despite being the quickest model to train, evaluate, and optimize. This model tends to predict that a higher ratio of examples does not contain either directed or guided cues. For example, the phrase "who manages the medications" is a guided cue in the validation data set, but the LSTM model predicted this to not be a cue. Interestingly, there are 6 phrases with the "who" trigger word within the training data set; 1 of these phrases is not a cue and the other 5 are guided cues. With this knowledge, it would make more sense for this example to have been predicted as a guided cue by the model. Similarly, the phrase "look over it for another minute or two" is a directed cue, yet the LSTM model predicted this to not be a cue. Within our training data set, we have 6 instances of phrases with the trigger word "look": 2 are not cues and the remaining 4 are directed cues. To a human, this example is easily identifiable as a command which would be considered a directed verbal cue from the strategy training coding scheme; the LSTM model, however, struggles to make this distinction.

The LSTM model also favored directed cues over guided cues; more specifically, this model had a higher number of false positive predictions for directed cues when compared to guided cues. Take as example the phrase "talk to me about the rules of cornhole": since this statement elicits thinking and memory skills, we would annotate it as a guided cue. While there are 2 guided cues in the training set that begin with the trigger words "talk to me," the LSTM model predicted this phrase to be a directed cue. In addition, the phrase "anything else you should watch out for in the floor" is labeled as a guided cue, but this model predicted it to be a directed cue. Like the previous example, there are 7 different examples of phrases that begin with the trigger words "anything else" in the training set, all of which are guided cues. There are 4 other instances of guided cues beginning with those same trigger words in the validation data set, all of which were incorrectly predicted to be directed cues by the LSTM model. The phrase "is that something you want to work on today" is the most interesting example of this phenomenon because there are 3 very similar guided cues in the training data set (e.g., "is that something you would want to work on"); despite this, the LSTM model predicted this case to be a directed cue.

Lastly, although the BERT model achieved the best results, there are noteworthy errors that the model made that could give insight on how to improve this model in the future. The first of these errors is a collection of 12 phrases beginning with the trigger words "how many" (e.g., "how many nickels are in a quarter," "how many quarters are in a dollar," etc.) labeled as not a cue that were predicted by the BERT model to be guided cues. Interestingly, there was a total of 3 phrases in the training data set beginning with the trigger words "how many": 2 of these phrases were not cues, and the remaining 1 was a guided cue. Further, the model correctly predicted that 2 similar "how many" phrases were not cues (e.g., "how many nickels are and a half dollar" and "how many 50 dollar bills are in 100"); therefore, it's curious that the model seemed to gravitate towards predicting guided cues for these phrases when the training data would indicate a stronger bias towards these phrases not containing cues. One potential explanation is that these 12 phrases all share a very similar sentence structure (e.g., "how many ____ are in a ____") and the reason the 2 similar phrases were predicted accurately by the BERT model could be that they differed ever so slightly in their sentence structure from the rest of the "how many" phrases.

Another interesting error that the BERT model made was its prediction for the phrase "what are some ways mhm could you tell." The model predicted this to not be a cue, when it's label was a guided cue. Out of the 76 phrases total, there were only 11 other phrases with the trigger word "what" that the BERT model predicted incorrectly: 3 of these were guided cues predicted to not be cues (like the example in question), 7 of them were not cues predicted to be guided cues, and the last 1 was a directed cue predicted to not be a cue. The interesting aspect about the phrase "what are some ways mhm could you tell" is that it contains the interjection "mhm," which is itself a trigger word in both the training and validation data sets. The vast majority of phrases that contain "mhm" are not cues of any kind. This detail might explain why the phrase "what are some way mhm could you tell" in particular would be predicted to not be a cue by the BERT model. Furthermore, when the word "mhm" is removed from all of the expressions during preprocessing, we observed a slight decrease of 0.02 in the F1 score of our model. It is therefore possible that "mhm" is an important key word for identifying none cues.

One more interesting error made by the BERT model occurred when the model classified the phrase "and then I'm going to take it away we'll see how many you remember." This phrase is a guided cue that was predicted by the model to be a directed cue. What makes this example noteworthy is that this phrase is one of the more difficult phrases to classify: this cue could be mistakenly labeled as directed because the therapist is telling the client what action they are planning to take next (e.g., "I'm going to take it away"), when the most important part of the phrase is "we'll see how many you remember" because it prompts the client to begin to devise a plan to help them remember what it is that they're working on. There were 12 similar phrases in the validation set; 9 of these were correctly predicted (2 were not cues, the remaining 7 were directed cues) by the BERT model and 3, including the current example, were predicted incorrectly. Of these 3, 2 were guided cues predicted to be directed and the other a directed cue predicted to not be a cue. What might make predicting this phrase difficult for the model is the fact that the data that it was trained on contains 45 phrases that begin with the trigger words "and then," of which 7 are not cues, 7 are guided cues, and 31 are directed cues. This may indicate a bias towards directed cues for "and then" phrases by this model.

Finally, we verified the performance of our BERT model using an external validation dataset collected from a separate major regional health system. The model was able to achieve an F1 score of **0.8259**, which exceeds the performance on our test dataset. This shows that the BERT model generalizes well when introduced to new data. The testing dataset also gives further insight into the errors or trends of our BERT model. For instance, there are 32 expressions in the testing dataset with the word "mhm" and each of these phrases is labeled as a none cue. The model was able to predict each of these none cues accurately, which may enforce the idea that "mhm" is an important key word for identifying none cues. Additionally, we found that the model mislabeled the direct category the least (i.e., 66 guided cues, 32 direct cues, and 58 none cues were mislabeled). However, we also found that, out of the wrong predictions, the model predicted the direct label the most (i.e., 39 phrases predicted to be guided, 85 to be direct, and 32 to be none). This may indicate that the model has a slight bias towards direct cues, since over half of the errors made by the model are attributed to the model incorrectly predicting a cue to be a direct cue.

**Discussion**
It is a challenging task to identify guided and directed verbal cues from strategy training of inpatient rehabilitation sessions. This project was the first to test an automated method for assessing fidelity of strategy training, or any rehabilitation intervention, using a published standardized fidelity assessment. Most promising is the fact that the automated method achieved industry standards for reliability in "real world settings" when compared with our trained evaluator ratings (≥.70), and simultaneously addressed previous limitations in labor, time, and cost. While we believe that the automated method can be refined further to achieve a higher reliability with the manual method, these preliminary findings are highly promising. Another future direction is to incorporate the video and audio information and to develop multimodal machine learning models[19-21] to increase the performance of identifying guided and directed cues.

There were several new insights brought forth by this study. To achieve a level of precision in annotation to support the automated method, the research team further refined the standardized fidelity assessment. In addition, when training non-rehabilitation personnel to annotate videos to train the algorithm, we became aware that the context of activities performed by patients and the interactions with therapists needed to be incorporated into the fidelity assessment. By doing this, we further advanced the reliability of the fidelity assessment, making the standardized tool more generalizable for use outside its original purpose.

That said, additional development is needed. In some cases, therapists used gestures or actions while talking with their patients. These visual cues can be categorized as directed and guided as well. In the next step of our work, we will analyze the video along with the audio to determine directed and guided gestures. We will start with the annotation of the collected videos. After we annotate the same set of videos used for the NLP work, we will use convolutional neural networks and semi-supervised learning algorithms to perform training and testing on the video data. Early focus will be on some major actions of therapists, such as demonstration of certain rehabilitation activities. The preliminary results will help us to determine whether we need to annotate more videos for the training to obtain more accurate results. In the future, we will also work on the recognition of fine motor activities performed by the therapists, such as moving a client's hand, pointing to an object, or demonstrating an action.

The findings from this study hold widespread promise in psychology and rehabilitation intervention research and practice. As we seek to characterize complex interventions in these settings for the purposes of standardization and fidelity assessment, the methods in this study can influence how we might leverage other well-characterized interventions for the use of automated fidelity assessment.

One limitation of this study is that we had a small amount of data compared to the vast, undocumented corpus of verbal cues that are used in-practice, though the total number of guided and directed verbal cues is not small. An extension to this is that the number of therapists involved in the rehabilitation sessions was relatively low. Different therapists may have various preference of using distinct words during the training with patients, which may impact the performance of NLP methods. Ideally, we would want to consider more therapy sessions and therapists when building the fidelity assessment lexicon to increase the variety of language used in guided and directed verbal cues. Further, adding more sessions with a larger number of therapists into the lexicon would diversify the language used since we hypothesize that each therapist has a unique phrasing technique that differs from other therapists (i.e., each therapist uses different words or phrasing when giving instructional cues).

One area of interest that may have an effect on the performance of each model is the ASR tool used to produce the transcripts. As previously mentioned, we used Amazon Web Service's ASR tool to produce the transcripts of this study; however, OpenAI has recently released an open source ASR tool called Whisper. This software has multiple model sizes that provide different levels of granularity in the transcribed audio, which may result in a more accurate transcription than our current tool. While we would like to investigate the performance of this software compared to our current tool, this would require another round of rigorous annotation of transcripts which is beyond the current scope of this project.

**Conclusion**
Strategy training is an efficacious multidisciplinary rehabilitation approach that shows promise for reducing disparities in patient and health system rehabilitation outcomes for people with cognitive impairments. The proposed NLP system showed promising results for automating the fidelity assessment by identifying guided verbal cues from recordings of rehabilitation sessions. Our study shows that using AI technologies could automate the fidelity assessment and facilitate the implementation of strategy training. The proposed technology could overcome the shortage of previous labor intensive, time-consuming, and expensive assessment, and to provide great solution to scale up future large clinical trials of meta-cognitive strategy training.

**Acknowledgement**
The authors would like to acknowledge the support from the National Institutes of Health through Grant Numbers UL1TR001857 and U24TR004111, the University of Pittsburgh Momentum Funds, and the School of Health and Rehabilitation Sciences Dean's Research and Development Award.

**References**
1. Skidmore ER, Dawson DR, Butters MA, et al. Strategy training shows promise for addressing disability in the first 6 months after stroke. *Neurorehabilitation and neural repair*. 2015;29(7):668-676.
2. Skidmore ER, Holm MB, Whyte EM, Dew MA, Dawson D, Becker JT. The feasibility of meta-cognitive strategy training in acute inpatient stroke rehabilitation: case report. *Neuropsychological rehabilitation*. 2011;21(2):208-223.
3. Schmidt R, Lee T. *Motor learning and performance 6th edition with web study guide-loose-leaf edition: From principles to application*. Human Kinetics Publishers; 2019.
4. Skidmore ER, Butters M, Whyte E, Grattan E, Shen J, Terhorst L. Guided training relative to direct skill training for individuals with cognitive impairments after stroke: a pilot randomized trial. *Arch Phys Med Rehab*. 2017;98(4):673-680.
5. Urquhart JR, Skidmore ER. Guided and directed cues: developing a standardized coding scheme for clinical practice. *OTJR: occupation, participation and health*. 2014;34(4):202-208.
6. Rouch S, Skidmore ER. Examining guided and directed cues in strategy training and usual rehabilitation. *OTJR: occupation, participation and health*. 2018;38(3):151-156.
7. Van Stan JH, Whyte J, Duffy JR, et al. Rehabilitation Treatment Specification System: Methodology to identify and describe unique targets and ingredients. *Arch Phys Med Rehab*. 2021;102(3):521-531.
8. Briganti G, Le Moine O. Artificial intelligence in medicine: today and tomorrow. *Front Med-Lausanne*. 2020;7:27.
9. Wang F, Casalino LP, Khullar D. Deep learning in medicine—promise, progress, and challenges. *JAMA internal medicine*. 2019;179(3):293-294.

10. Piccialli F, Di Somma V, Giampaolo F, Cuomo S, Fortino G. A survey on deep learning in medicine: Why, how and when? *Information Fusion*. 2021;66:111-137.
11. Benzeghiba M, De Mori R, Deroo O, et al. Automatic speech recognition and speech variability: A review. *Speech communication*. 2007;49(10-11):763-786.
12. Wang D, Wang X, Lv S. An overview of end-to-end automatic speech recognition. *Symmetry*. 2019;11(8):1018.
13. Li J. Recent advances in end-to-end automatic speech recognition. *APSIPA Transactions on Signal and Information Processing*. 2022;11(1)
14. Wang Y, Wang L, Rastegar-Mojarad M, et al. Clinical information extraction applications: a literature review. *Journal of biomedical informatics*. 2018;77:34-49.
15. Sorin V, Barash Y, Konen E, Klang E. Deep learning for natural language processing in radiology—fundamentals and a systematic review. *Journal of the American College of Radiology*. 2020;17(5):639-648.
16. Otter DW, Medina JR, Kalita JK. A survey of the usages of deep learning for natural language processing. *IEEE transactions on neural networks and learning systems*. 2020;32(2):604-624.
17. Yeung A, Iaboni A, Rochon E, et al. Correlating natural language processing and automated speech analysis with clinician assessment to quantify speech-language changes in mild cognitive impairment and Alzheimer's dementia. *Alzheimer's research & therapy*. 2021;13(1):1-10.
18. Carson NJ, Mullin B, Sanchez MJ, et al. Identification of suicidal behavior among psychiatrically hospitalized adolescents using natural language processing and machine learning of electronic health records. *PloS one*. 2019;14(2):e0211116.
19. Baltrušaitis T, Ahuja C, Morency L-P. Multimodal machine learning: A survey and taxonomy. *IEEE transactions on pattern analysis and machine intelligence*. 2018;41(2):423-443.
20. Baltrušaitis T, Ahuja C, Morency L-P. Challenges and applications in multimodal machine learning. *The Handbook of Multimodal-Multisensor Interfaces: Signal Processing, Architectures, and Detection of Emotion and Cognition-Volume 2*. 2018:17-48.
21. Xu K, Lam M, Pang J, et al. Multimodal machine learning for automated ICD coding. PMLR; 2019:197-215.